# OncoGPT: A Medical Conversational Model Tailored with Oncology Domain Expertise on a Large Language Model Meta-AI (LLaMA)

Fujian Jia[1,*], Xin Liu[1,*,§], Lixi Deng[1,*], Jiwen Gu[1], Chunchao Pu[1], Tunan Bai[1], Mengjiang Huang[2], Yuanzhi Lu[3,§], Kang Liu[1,§]

1, Shenzhen Kanghua Juntai Biotech Co. Ltd., B 215, Unit No.7, Shahe Rd W, Nanshan, Shenzhen, Guangdong Province 518063, China.

2, Department of Nutrition and Graduate Group in Nutritional Biology, University of California, Davis, CA 95616, USA.

3, Department of Pathology, The First Affiliated Hospital of Jinan University, Tianhe Qu, Guangzhou 510632, China.

*, Equal contribution

§, Correspondence:

Xin Liu, xin_liu@kanghuajuntai.com

Yuanzhi Lu, Yuanzhi.lu@jnu.edu.cn

Kang Liu, kellen0101@live.com

# Abstract

In the past year, there has been a growing trend in applying Large Language Models (LLMs) to the field of medicine, particularly with the advent of advanced language models such as *ChatGPT* developed by OpenAI. However, there is limited research on LLMs specifically addressing oncology-related queries. The primary aim of this research was to develop a specialized language model that demonstrates improved accuracy in providing advice related to oncology. We performed an extensive data collection of online question-answer interactions centered around oncology, sourced from reputable doctor-patient platforms. Following data cleaning and anonymization, a dataset comprising over 180K+ oncology-related conversations was established. The conversations were categorized and meticulously reviewed by field specialists and clinicians to ensure precision. Employing the *LLaMA* model and other selected open-source datasets, we conducted iterative fine-tuning to enhance the model's proficiency in basic medical conversation and specialized oncology knowledge. We observed a substantial enhancement in the model's understanding of genuine patient inquiries and its reliability in offering oncology-related advice through the utilization of real online question-answer interactions in the fine-tuning process. We release database and models to the research community (https://github.com/OncoGPT1).

# Introduction

The advent of *ChatGPT* has brought unprecedented attention to instruction-following LLMs have gained unprecedented attention, owning to their remarkable ability to comprehend human queries and generate responses in a human-like manner[1]. These autoregressive LLMs demonstrate strong robustness and generalization capabilities in natural language processing tasks. By training on a large set of dialogues, they learn the underlying logic of human conversations and continuously generate the most likely next token based on the preceding context, resulting in realistic human-like responses[2]. Despite ChatGPT not currently being open source, several LLMs such as *LLaMA* and *BERT* are available in open-source communities[3, 4]. Leveraging these accessible LLMs and fine-tuning them with specific training data of interest can provide personalized and potentially cost-effective solutions for a wide range of practical communication and reasoning needs.

While LLMs possess certain capabilities in answering medical domain questions and demonstrate the potential to transform medical communication, their current general knowledge is not specifically tailored to the medical field. They lack in-depth understanding of specialized medical expertise, resulting in frequent inaccuracies in their responses to medical queries. While various medical-related large language models are emerging, Google's *Med-PaLM* is a noteworthy LLM designed to deliver high-quality answers to medical questions[5]. Chatbots like *First Derm* and *Pahola* are already aiding doctors in assessing and guiding patients with skin conditions and alcohol abuse[6]. However, these chatbots may require further enhancements in functionality and acceptance by medical professionals. Models specifically addressing oncology question-answer dialogues are rarely reported, primarily due to the scarcity of high-quality specialty datasets related to cancer.

In this study, we focused on systematically collecting, cleaning and proof-reviewing to establish a high-quality oncology-related dialogue data between professional doctors and patients. Since the prevalent *ChatGPT* model is not open source, we used Meta's popular public *LLaMA-7B* model as the platform for development and evaluation. By training the *LLaMA-7B* base model with 52K instruction data from Stanford University's publicly available *Alpaca* model[7] and incorporating 100K patient-physician conversations from an online medical consultation website (from *ChatDoctor* LLM)[8], the model successfully acquired general medical conversation abilities. Subsequently, the model underwent further refinement using our established 180K+ real-world doctor-patient oncology dialogue dataset, resulting in the creation of OncoGPT, our specialized LLM model tailored for oncology-related queries. When compared to other LLM medical models like *ChatDoctor*, OncoGPT demonstrates a notable improvement in understanding the needs of cancer patients and effectively addressing cancer-related inquiries.

In sum, our work principally contributes to the following:

1. We created and openly shared a high-quality dataset consisting of

oncology-related patient-doctor conversations, serving as a valuable resource for refining medical LLMs. This dataset underwent thorough proof-review by field specialists and clinicians, incorporating a diverse range of specialized medical knowledge relevant to oncology.

2. We designed the OncoGPT v1.0 LLM specifically for professional question-answer sessions related to online oncology consultations. This model offers a relatively dependable and efficient consultation channel for oncology patients. The OncoGPT training data and related codes can be accessed at https://github.com/OncoGPT1.

## Materials And Methods

**Establishment of oncology-specific patient-doctor conversation dataset**

To closely emulate the actual question-answer scenarios of oncology patients, we first collected all medical-related conversation data from online medical consultation websites, which more closely resemble actual chat formats, for the model's fine-tuning input data. These data primarily comprised 2 parts: 1) English patient-doctor dialogues mostly retrieving from HealthCareMagic, eHealth Forum, Question Doctors, WebMD, among other sites. 2) Chinese patient-doctor dialogues from websites including https://www.youlai.cn/, https://www.miaoshou.net/, https://www.mfk.com/, https://www.bohe.cn/, https://www.iplusmed.com/, https://www.yilianmeiti.com/, https://wwys.familydoctor.com.cn/,

https://www.xywy.com/. The Chinese conversations were batch-translated into English using the *Google Translation* tool and meticulously reviewed by human.

Subsequently, we utilized keyword filtering to identify conversations related to cancer. This involved selecting dialogues containing keywords such as cancer, carcino-, tumor, carcinoma, neoplasm, malignancy, lesion, leukemia, adenoma, and sarcoma, ensuring the retention of dialogue information pertinent to oncology. To enhance data reliability, we conducted thorough cleaning of the obtained dialogue data, which included the removal of question-answer dialogues that were irrelevant to the posed questions. In summary, a total of 180,184 dialogues related to oncology were extracted. Detailed information is provided in Table 1.

**Table 1. Source of Oncology-specific Dialog Data Collected**

| Source Name | Language | # of dialogues | Link |
| --- | --- | --- | --- |
| MiaoShou | Chinese | 93 | https://www.miaoshou.net/ |
| YouLai | Chinese | 120 | https://www.youlai.cn/ |
| MFK | Chinese | 550 | https://www.mfk.com/ |
| BoHe | Chinese | 327 | https://www.bohe.cn/ |
| XYWY | Chinese | 1854 | https://www.xywy.com/ |
| YiLianMeiTi | Chinese | 2343 | https://www.yilianmeiti.com/ |
| FamilyDoctor | Chinese | 3141 | https://wwys.familydoctor.com.cn/ |
| iPlusMed | Chinese | 149799 | https://www.iplusmed.com/ |
| eHealth Forum | English | 1159 | https://www.iehealthf.com/ |
| iCliniq | English | 735 | https://www.icliniq.com/ |
| Question Doctors | English | 1271 | https://questiondoctors.com/ |
| WebMD | English | 4253 | https://www.webmd.com/ |
| HealthCareMagic | English | 14539 | https://www.healthcaremagic.com/ |

To protect privacy and mitigate the influence of redundant, non-critical information on the model's effectiveness during subsequent fine-tuning, we took measures to exclude names, hospitals, and website information from the dialogues. This step ensures that all data is devoid of personal information, adhering strictly to privacy protection and ethical standards. Subsequently, we categorized the dialogues into fundamental-related questions (pertaining to popular science) and treatment-related questions. The data underwent review by corresponding field specialists and clinicians to rectify inaccuracies and ensure the authority of the answers. In sum, we successfully created a high-quality, proof-reviewed oncology-related dataset consisting of 180K+ patient-doctor conversations, as illustrated in Figure 1.

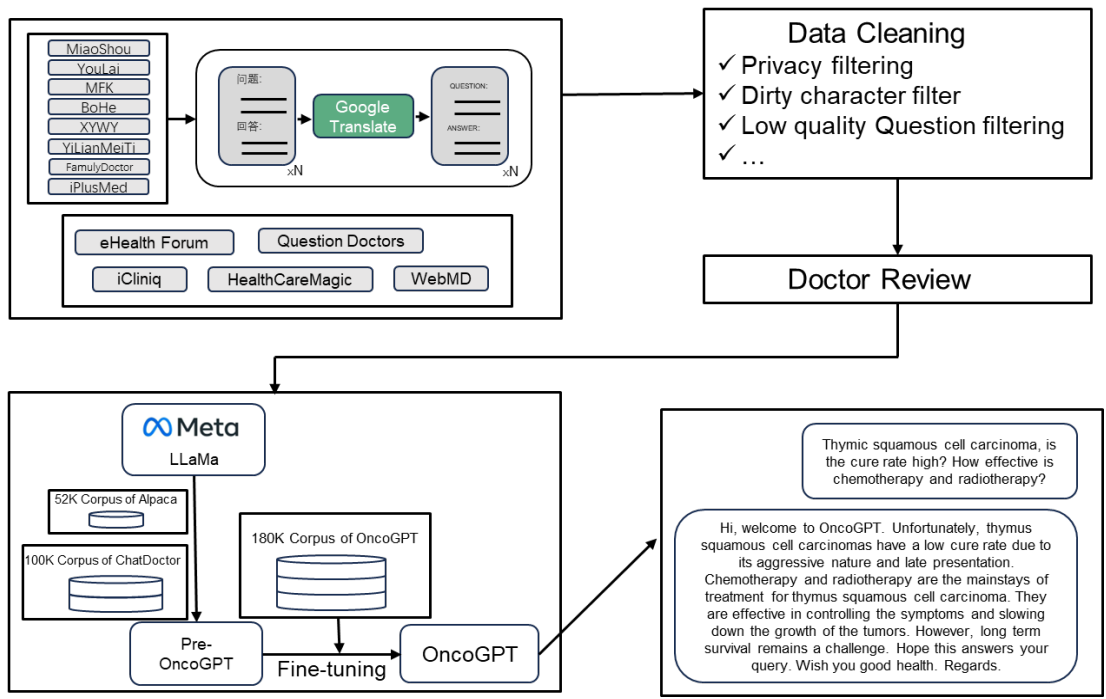

**Figure1.** Overview of the physician and patient conversation dataset collection pipeline and the training procedure of OncoGPT.

**Model Training**

Meta's open-source *LLaMA-7B* model, comprised of 7 billion parameters, has demonstrated performance comparable to the larger-scale *GPT-3.5* model across multiple natural language processing benchmark tests. It is currently extensively employed across various domains to generate dialogue models for specialized applications. Similar to the Alpaca model, our model was initially fine-tuned using a dataset of 52K to acquire basic dialogue capabilities. It was then fine-tuned using dialogues generated from the 5K disease database from *ChatDoctor*, enabling the model to learn the intrinsic knowledge relating diseases to symptoms, medical tests, and recommended medications. Further refinement involved fine-tuning with a 100K dataset to construct a medical question-answer model resembling *ChatDoctor*. Lastly, fine-tuning was carried out using our established 180K+ authentic oncology-related dialogue datasets. This fine-tuning process on the 180K+ authentic doctor-patient dialogue datasets utilized 1*A100 GPUs over a duration of 15 hours. This was followed by a series of parameter optimization steps, culminating in the development of our OncoGPT model.

**Optimal Parameter Selection**

Through a systematic manipulation of a range of parameters, including those linked to Lora and Llama, we embarked on an iterative process of refinement. This entailed a series of experiments to determine the optimal parameter configuration, further enhancing the model's performance.

Considering the limitations posed by data volume and training time costs, we initially utilized a subset of 100K data samples for constructing the training model. We then employed the aforementioned set of 350 test data, following the established criteria, to evaluate the similarity between responses generated by multiple trained models under various parameter selections and actual doctor-patient responses. This comparative analysis aimed to identify the model yielding the most optimal responses. Subsequently, the parameters of this selected model were chosen as the default configuration for the subsequent training of the OncoGPT model. The hyperparameters employed in the training process were as follows: the total batch size of 128, a learning rate of $3\times10^{-4}$, a total of 3 epochs, a maximum sequence length of 512 tokens.

The testing iterations rigorously aligned model outcomes with authentic medical professional responses, employing statistical metrics like correlation, recall, and precision for quantitative assessment. The overarching goal was to identify paramount parameters enhancing output accuracy, ensuring model-generated responses closely resemble those of real doctors, with heightened precision and nuanced knowledge.

**Model Validation**

To ensure a more precise evaluation of OncoGPT performance, a thorough human-led selection process was employed to extract a subset of 737 dialogues from the extensive 180K+ oncology conversation datasets. The primary criterion for curation was to prioritize conversations focusing on fundamental oncology knowledge and therapeutic interventions. Notably, during the training phase of the OncoGPT model, these 737 curated dialogue instances were intentionally withheld and not included in the 180,184 datasets used for training. This specific subset of dialogues was deliberately preserved to function as an external validation set for subsequent model assessments.

The responses generated by OncoGPT and *ChatDoctor* underwent a systematic comparison with authentic medical professional responses, facilitating a meticulous examination of their congruence. The primary aim of this evaluation was to determine the extent to which OncoGPT's responses aligned with genuine medical expertise, contrasting with the responses generated by *ChatDoctor*.

To assess the quality of generated text, particularly in the context of natural language processing, the BERT Score method was employed. BERT Score involves comparing embeddings between the generated text and reference text, calculating cosine similarity, and computing Precision, Recall, and F1 Score based on this similarity. The Precision, Recall, and F1 scores of *ChatDoctor* and OncoGPT responses, compared to the doctor's responses, were determined using BERT scores, respectively. This comprehensive analysis, facilitated by BERT Score, allowed for a comparative evaluation of the performance of OncoGPT and *ChatDoctor*.

## Result

**OncoGPT Performance**

To evaluate the performance of the OncoGPT model effectively, we used the previously mentioned randomly selected 737 authentic patient inquiries as inputs. Simultaneously, identical inputs were provided to the *ChatDoctor* model, and their responses were recorded. BERT scoring was then applied to calculate accuracy, recall, and F1 scores for OncoGPT and *ChatDoctor*. This approach facilitated a comprehensive and objective assessment of the models' performance.

As illustrated in Figure 2, it is evident that the OncoGPT model, further fine-tuned using our 180K+ oncology data, outperformed both the *ChatDoctor* in terms of precision, recall, and F1 scores.

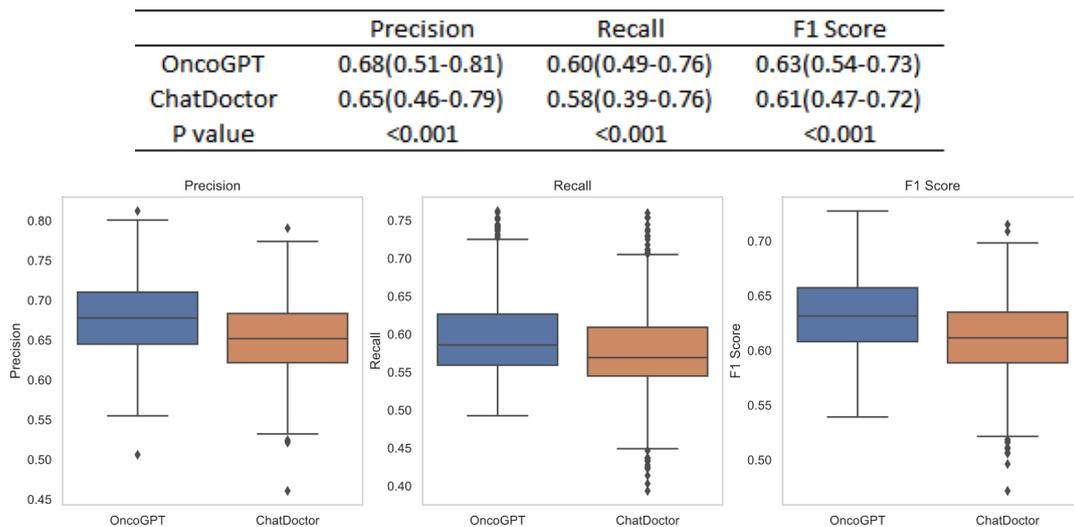

**Figure 2.** Overall Model Performance Evaluation

Additionally, we conducted a comparative performance test of the OncoGPT model in two distinct categories of questions: fundamental-related and treatment-related. In both categories, the OncoGPT model exhibited superior performance compared to

*ChatDoctor* (Figure 3 and 4).

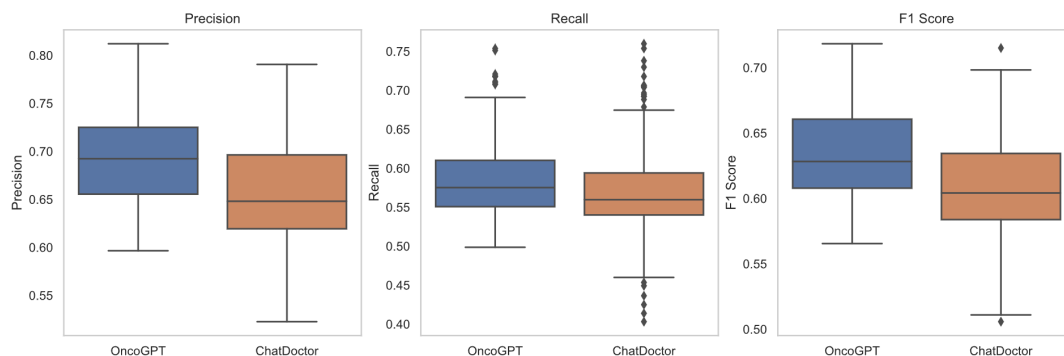

|  | Precision | Recall | F1 Score |
| --- | --- | --- | --- |
| OncoGPT | 0.69(0.60-0.81) | 0.58(0.50-0.75) | 0.63(0.57-0.72) |
| ChatDoctor | 0.65(0.52-0.79) | 0.56(0.40-0.76) | 0.60(0.50-0.72) |
| P value | <0.001 | <0.01 | <0.001 |

**Example**

**Q:** How effective is apatinib in the treatment of lung cancer?

**OncoGPT:** Hi, Thanks for writing in. Apatinib is a tyrosine kinase inhibitor used in the treatment of non-small cell lung cancer. It is claimed to improve survival in patients with advanced or metastatic NSCLC who have received prior chemotherapy and have an activating mutation in the EGFR gene. However, the effectiveness of this drug is still being studied and its long term effects are yet to be seen.

**ChatDoctor:** Hi, Thanks for writing in. Apatinib is a targeted therapy for an activating mutation in the EGFR gene. It is being studied in clinical trials for the treatment of certain types of lung cancer but its effectiveness is unclear. There is currently no way to guarantee a cure for any type of cancer, including lung cancer. Wish you good health.

**Figure 3.** Model Performance Evaluation in Fundamental-related Oncology Questions

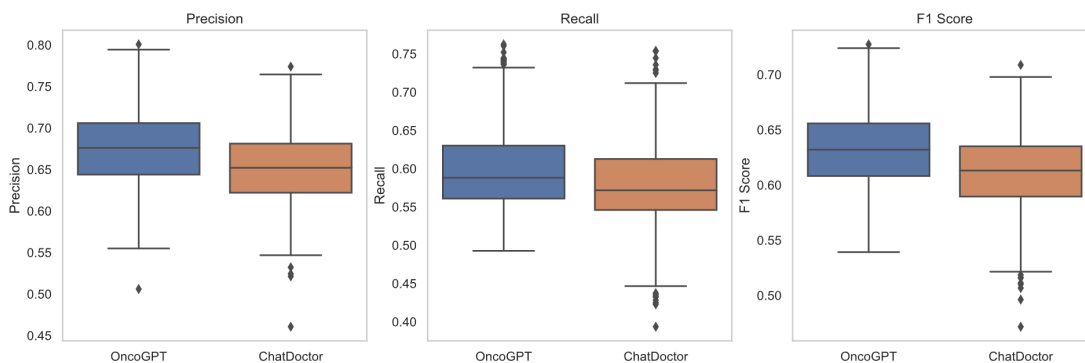

|  | Precision | Recall | F1 Score |
|---|---|---|---|
| OncoGPT | 0.68(0.51-0.80) | 0.59(0.49-0.76) | 0.63(0.54-0.73) |
| ChatDoctor | 0.65(0.46-0.77) | 0.57(0.39-0.75) | 0.61(0.47-0.71) |
| P value | <0.001 | <0.001 | <0.001 |

**Example**

**Q:** Lung metastases and rib metastases after breast cancer surgery, is there still hope? The patient is currently emaciated, coughing, with profuse phlegm in the form of foam, one side of the lung is enlarged, and chemotherapy has no effect. May I ask whether to continue chemotherapy or to take traditional Chinese medicine? Is there any hope? Thank you so much sir!

**OncoGPT:** Thanks for posting your query. Yes, there is still hope. Chemotherapy is not always effective in treating metastatic disease. In such cases, we usually recommend other treatment options such as targeted therapies, immunotherapy, and traditional Chinese medicines. These are more effective in controlling the symptoms and slowing down the progression of the disease. We also recommend nutrition supplementation and physical exercise to improve the general condition of the patient. Please do not hesitate to ask us further questions if any. Wish you good health and early recovery. Best regards.

**ChatDoctor:** Hi, Thanks for writing in. Yes, there is still hope. Chemotherapy can help prolong survival but not cure the disease. TCM can help in palliation and improve quality of life. Please consult a TCM doctor and discuss further options. Wishing you good health. Take care.

**Figure 4.** Model Performance Evaluation in Treatment-related Oncology Questions

## Discussion

The evolution of LLMs has given rise to numerous solid, open-source frameworks like *LLaMA[3]*, *Vacuna[9]* and *BERT[4]*. Emphasizing the significance of computational power and datasets to deliver high-quality domain-specialized LLM, it is evident that particularly in the healthcare area, the availability of high-quality open medical training datasets is limited. Previous research on the medical LLM called "*ChatDoctor*" indicates the possibility of enhancing both the overall accuracy and mitigating "hallucinations" in medical LLMs, even compared to *ChatGPT 3.5* in the medical context[8]. This improvement is achieved through the fine-tuning of the "lightweight" LLaMA-7B model, using exclusively online-retrieved real-world patient-doctor interactions.

In current study, we developed OncoGPT as a medical LLM chatbot by further fine-tuning a *LLaMA-7B* model using a real oncology-related doctor-patient dialogue database. It is primarily designed to address cancer-related queries and holds the potential for a variety of applications. To our knowledge, this is the first model/datasets constructed for oncology-specific LLM. As the prevalence of cancer continues to rise, there is an increasing need for accessible and more precise expert guidance on cancer-related matters.

Our primary contribution lies in the creation of a large high-quality, expert-proof-reviewed, open-source medical dialogue database (180K+ patient-doctor conversations) with a specific emphasis on oncology-related questions. Furthermore, we demonstrate that by incorporating conversation data in languages other than English, using a batch-translation script, the "translated" datasets substantially enhance overall quality without encountering noticeable issues. This suggests the potential for adding more data sources in the future to further augment field-specific datasets. It's noteworthy that the current dataset also includes a substantial number of conversations related to cancer therapy/recovery in traditional Chinese medicine, largely contributed by Chinese online sources. As expected, by leveraging this database, OncoGPT demonstrated superior performance in addressing oncology-related questions compared to ChatDoctor.

It is important to highlight that the current OncoGPT model is still in the investigational stage and should exclusively be utilized for academic research purposes. Additional functionalities, such as automated reference checking and citation (similar to OpenEvidence, Mayo Clinic Platform Accelerate Company, 2023), and the implementation of Reinforcement Learning from Human Feedback (RLHF)[10], are deemed necessary to enhance the model. These enhancements are crucial for identifying potentially inaccurate answers and preventing hallucinations. Furthermore, training the model with more advanced frameworks, such as *LLaMA2*, has the potential to further improve overall quality and ensure consistency in the models.

In sum, OncoGPT exhibits significant potential for diverse oncology-related practical applications including professional consultation and educational services. Primarily, it may serve as an intelligent consultation service on medical platforms, it efficiently responds to queries, easing the workload of medical personnel. This alleviates communication challenges between doctors and patients, particularly in regions with limited healthcare resources. OncoGPT is poised to provide essential assistance to both patients and healthcare providers, thereby contributing to advancements in medical research.